%% file: d2-iac2023.tex
\documentclass[]{iac}

\usepackage[table]{xcolor}
\usepackage{svg}
\usepackage{subcaption}
\usepackage{placeins}
\usepackage{multirow}
\usepackage{wrapfig}
\usepackage{adjustbox}
\usepackage{textcomp}
\usepackage{gensymb}
\usepackage{array}
\usepackage{soul}
\usepackage{siunitx}

\usepackage{chemformula}

\usepackage[acronym, nonumberlist]{glossaries}
\makeglossaries
\input{abbreviations}

\usepackage[obeyFinal]{todonotes}
\presetkeys{todonotes}{inline}{}

\newcommand{\D}{} 
\protected\def\D #1{Daedalus~#1}

\IACpaperyear{23}
\IACpapernumber{E2,3-GTS.4,6,x78230}
\IACconference{74}{th}
\IAClocation{Baku, Azerbaijan}
\IACdate{2-6 October 2023}
\IACcopyrightB{2023}{the authors}

\title{Daedalus 2: Autorotation Entry, Descent and Landing Experiment on REXUS29}

\IACkeyword{Autorotation, Landing, Decelerator, Descent, EDL}

\IACauthor[true]{Philip Bergmann}{Julius-Maximilians-Universität Würzburg, Germany}{philip.bergmann@wuespace.de}
\IACauthor{Clemens Riegler}{Julius-Maximilians-Universität Würzburg, Germany}{clemens.riegler@uni-wuerzburg.de}
\IACauthor{Zuri Klaschka}{Julius-Maximilians-Universität Würzburg, Germany}{zuri.klaschka@wuespace.de}
\IACauthor{Tobias Herbst}{Julius-Maximilians-Universität Würzburg, Germany}{tobias.herbst@uni-wuerzburg.de}
\IACauthor{Jan M. Wolf}{Julius-Maximilians-Universität Würzburg, Germany}{jan.wolf@wuespace.de}
\IACauthor{Maximilian Reigl}{Julius-Maximilians-Universität Würzburg, Germany}{maximilian.reigl@wuespace.de}
\IACauthor{Niels Koch}{Julius-Maximilians-Universität Würzburg, Germany}{niels.koch@wuespace.de}
\IACauthor{Sarah Menninger}{Julius-Maximilians-Universität Würzburg, Germany}{sarah.menninger@wuespace.de}
\IACauthor{Jan von Pichowski}{Julius-Maximilians-Universität Würzburg, Germany}{jan.pichowski@wuespace.de}
\IACauthor{Cedric Bös}{Julius-Maximilians-Universität Würzburg, Germany}{cedric.boes@wuespace.de}
\IACauthor{Bence Barthó}{Julius-Maximilians-Universität Würzburg, Germany}{bence.bartho@wuespace.de}
\IACauthor{Frederik Dunschen}{Julius-Maximilians-Universität Würzburg, Germany}{frederik.dunschen@wuespace.de}
\IACauthor{Johanna Mehringer}{Julius-Maximilians-Universität Würzburg, Germany}{johanna.mehringer@wuespace.de}
\IACauthor{Ludwig Richter}{Julius-Maximilians-Universität Würzburg, Germany}{ludwig.richter@wuespace.de}
\IACauthor{Lennart Werner}{ETH Zürich, Switzerland}{lennart.werner@wuespace.de}

\abstract{%
In recent years, interplanetary exploration has gained significant momentum, leading to a focus on the development of launch vehicles.
However, the critical technology of \gls{edl} mechanisms has not received the same level of attention and remains less mature and capable.
To address this gap, we took advantage of the REXUS program to develop a pioneering \gls{edl} mechanism.
We propose an alternative to conventional, parachute based landing vehicles by utilizing autorotation.
Our approach enables future additions such as steerability, controllability, and the possibility of a soft landing.
To validate the technique and our specific implementation, we conducted a sounding rocket experiment on REXUS29.
The systems design is outlined with relevant design decisions and constraints, covering software, mechanics, electronics and control systems.
Furthermore, an emphasis will also be the organization and setup of the team entirely made up and executed by students.
The flight results on REXUS itself are presented, including the most important outcomes and possible reasons for mission failure.
We have not archived an autorotation based landing, but provide a reliable way of building and operating such vehicles.
Ultimately, future works and possibilities for improvements are outlined.
The research presented in this paper highlights the need for continued exploration and development of \gls{edl} mechanisms for future interplanetary missions.
By discussing our results, we hope to inspire further research in this area and contribute to the advancement of space exploration technology.
}

\begin{document}

\makeatletter
\lhead{}\chead{\footnotesize \iac@conference{} International Astronautical Congress (IAC), %
  {\iac@location, \iac@date}. \\ \iac@makecopyright}\rhead{}%
  \lfoot{IAC--\iac@paperyear--\iac@papernumber}\cfoot{}\rfoot{Page \thepage\ of \pageref{LastPage}}%
\makeatother

\maketitle
\printglossaries

\section{Introduction}
Precise, soft landings remain an unsolved problem for planetary exploration and sample return.
There are already multiple technologies in use today:
The most well-known method are parachutes, as used for example with the OSIRIS-REx \cite{reza2014overview}, Apollo \cite{ewing1972ringsail} or Orion \cite{taylor2007developing}.
Other well-known methods include propulsive assisted landing, which gained much publicity from the sky crane landing of Curiosity.
Its biggest advantages over parachutes are not requiring an atmosphere and enabling midair steering and control over the touchdown velocity.
However, the trade-offs are increased weight because of the need to carry fuel.

This paper introduces a different \gls{edl} scheme, also used in terrestrial aviation:
Autorotation is used by helicopter pilots to safely land the aircraft after an engine failure.
It combines some of the advantages of both the parachute and the propulsive landing schemes,
namely requiring no fuel but theoretically enabling control of the landing site and velocity.
These techniques have also been investigated in the past, e.g. the KRC-6 by NASA \cite{robinson1963investigation} or ESA's ARMADA \cite{ARMADA} and AMDL \cite{AMDL} studies.

This paper presents a new mission that investigates autorotation for \gls{edl} called \D2.
We begin by introducing the mission and its components in \autoref{sec:mission-profile} before addressing some core design choices in more detail in \autoref{sec:exp-design}.
The flight, the mission failure and possible causes are detailed in \autoref{sec:flight}.
Finally, in \autoref{sec:team} we discuss the student team that designed and conducted the experiment as well as the organizational strategy and challenges that occurred during the project.

\section{Mission Profile}
\label{sec:mission-profile}
\D2 is a student project that was developed in the context of the REXUS/BEXUS program and flown on REXUS29.
The project organizers (DLR/SNSA/ESA) provide REXUS student teams with launch opportunities aboard two sounding rockets per cycle.
The rocket platform provides power and a basic telemetry link for the experiments on board.
A cycle typically takes 1.5 years (for the actual duration of this project, see \ref{sec:project-timeline}), during which the teams get assistance from DLR, SSC, ZARM and ESA experts while designing and building their experiment.

\subsection{System Topology}  
\begin{figure*}
    \centering
    \includegraphics[width=\textwidth]{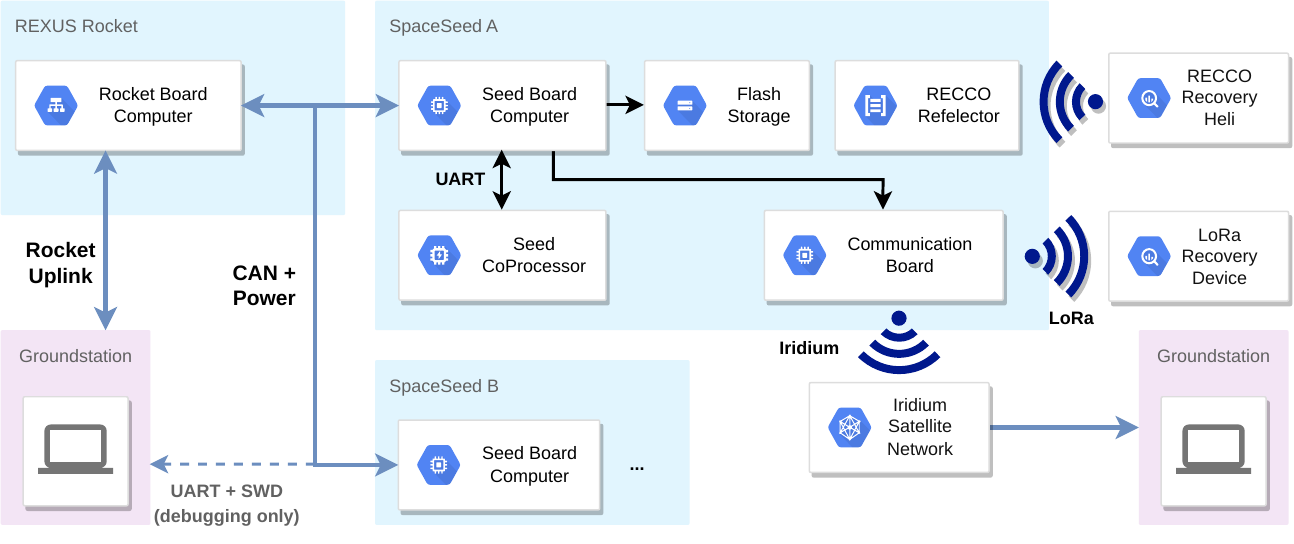}
    \caption{Overview of the \D2 system topology. Adapted from \cite{wolf2022electronics}.}
    \label{fig:exp-overview}
\end{figure*}

A schematic overview of the flight segment and parts of the ground segment of our experiment is depicted in \autoref{fig:exp-overview}.
The ground segment consists mainly of two instances of the ground station software with different configurations: one air-gaped deployment which interacts with the rocket and another internet-connected installation, which receives messages from the Iridium network.
The user segment, which consists mainly of data analysis using MATLAB and Python and is not in scope of this publication.

\begin{figure}
    \centering
    \includegraphics[width=0.96\linewidth]{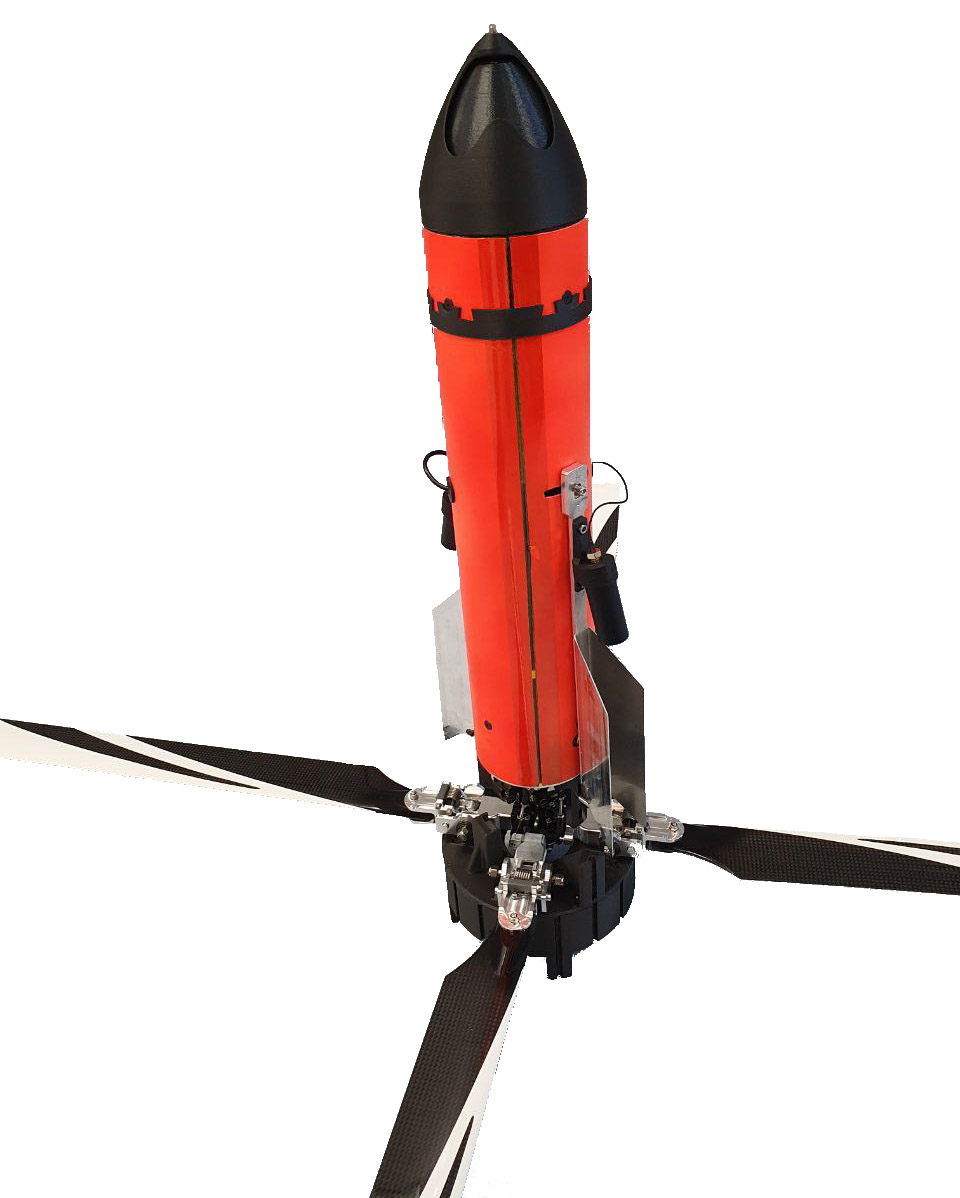}
    \caption{SpaceSeed with hull, upside-down with respect to falling direction.}
    \label{fig:seed-with-hull}
\end{figure}

The flight segment of our experiment consists of three parts:
Two \glspl{ffu} (the so-called SpaceSeeds) form the main part and are depicted in \autoref{fig:seed-with-hull}.
They are ejected at the trajectory apex and from that point on descend separately and autonomously with the goal of demonstrating autorotation, deceleration and landing.
At the end of the descent, the energy stored in the rotation of the rotor can be converted to lift to ensure a soft landing.
After separation, communication cannot be routed through rocket telemetry.
Power, telemetry and flight controls thus are handled independently by the units.
In the flown configuration both units were identical and provided shallow redundancy for increasing the probability of recovery.

The ejector is the third part of the flight segment.
It remains attached to the rocket and provides the mechanical support for the \glspl{ffu} during the ascent, and holds the mechanism used for ejection at apogee.
As the only rocket-interfacing part it also acts as the router for telemetry and telecommands transmitted through the rocket support system to and from the SpaceSeeds during pre-launch ground operations.

The \glspl{ffu} store their telemetry on on-board flash storage, as live downlink does not support the required data rates nor guarantees success of reception. 
Therefore, in order to determine the experiment outcome, the high-rate logged data must be extracted from the SpaceSeeds after landing, requiring recovery of the units.
Redundant localization schemes are used to make successful recovery more probable.
The main scheme has the \glspl{ffu} transmitting their \gls{gnss} position via the Iridium network.
Iridium messages might only get transmitted from high altitudes or not at all, requiring a prediction of the landing location based on current wind data.
Additionally, each seed broadcasts their position via LoRa in case only a course position is available which can aid in such cases.
RECCO reflectors, which are search and rescue tags, normally integrated into clothing for locating avalanche victims, are also integrated in the hull.
They enable a suitably equipped helicopter to locate the units, should their course location be known.

\subsection{Flight profile}
\label{sec:profile}
\begin{figure}
    \centering
    \includegraphics[width=0.96\linewidth]{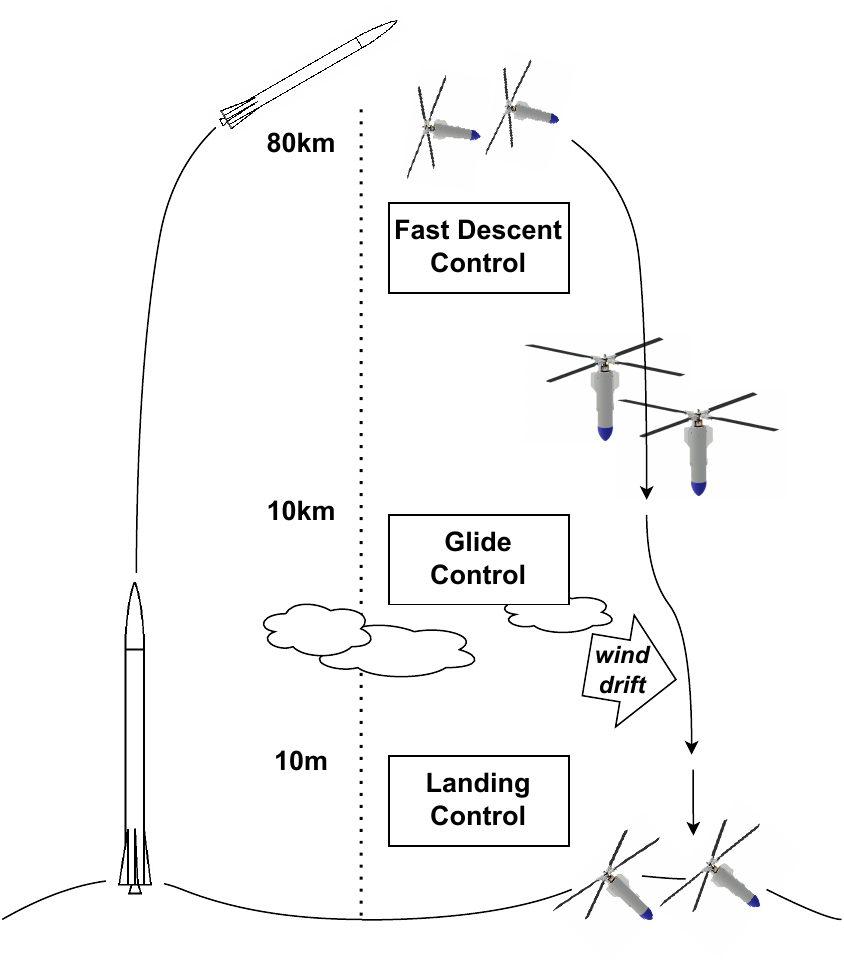}
    \caption{Flight Profile of Daedalus 2 \cite{riegler2022modeling}}
    \label{fig:profile}
\end{figure}

After nosecone separation, the \glspl{ffu} are ejected and their autonomous descent begins.
The flight profile can be divided into three critical zones that affect the control of the vehicle.
The Zones are visualized in \autoref{fig:profile} \cite{riegler2022modeling}.

\paragraph{Fast Descent}
In this phase, the vehicle is reentering the atmosphere from roughly \qty{75}{\km}.
Speeds of up to Mach 2 are expected in this phase.
During this time, the rotor is fully stalled and only acts as a drag surface.
Everything in this phase is geared towards stable flight and preventing damage to any parts of the vehicle.
Fast Descent can be left as soon as flight is possible. 
However, it might be continued to reach the surface faster.
Depending on the vehicle design, this is expected at around \qty{10}{\km} or below.
For \D2, this phase is kept until a height of \qty{2}{\km} above the surface is reached.
This was done to clear the airspace faster and minimize flight time to comply with launch site safety regulations.

\paragraph{Glide}
Once the fast descent phase ends, the glide can begin.
In this zone, the vehicle performs autorotation deceleration and flight.
\D2 prepares for landing during the glide phase, reducing vertical speed, increasing rotor rotation rate, and opening the LIDAR hatch to have a clear ground view.
Glide ended at \qty{100}{\m} above the surface.

\paragraph{Landing}
During landing, vehicle speed and ground distance is measured continuously.
At predefined conditions the SpaceSeed performs a flare maneuver with a target landing speed of \qty{0.5}{\m\per\s}.
Once landing was detected, the system switches into a recovery mode, transmitting location data alongside other telemetry.

\section{Experiment Design}
\label{sec:exp-design}
This section presents the flown design of our experiment.
The presented design is the product of an iterative design process guided by numerous tests as usual for student-lead engineering projects.
The COVID crisis required distributed work on different subsystems without continuous integration.
Rigorous integration tests guided the final work on each subsystem.

\subsection{Mechanics}

The mechanics consist of two interacting, but generally separate systems: The \glspl{ffu}s themselves and the ejector which stores them during launch and deploys them at a set time.
The first subsection will shortly outline the ejector design and the interactions it has with the SpaceSeeds, which will be discussed further on.
This chapter just gives a short overview of the key components, a more detailed description is available in \cite{mehringer2022suborbital}.

\subsubsection{Ejector}
The ejector is primarily composed of two tubes designed to securely hold the SpaceSeeds during their time aboard the rocket. These structural components are crafted using carbon-nylon filament through a 3D-printing process. Each of these tubes has an inner diameter measuring \qty{144}{\mm}. To attach it to the rocket, the 3D-printed structure is affixed to an aluminum interface plate.
Inside each tube, there are springs that get compressed when the \glspl{ffu} are inserted. After these units are loaded, the springs remain compressed, held in position by a steel wire. This wire is cut by pyrocutters when the rocket reaches the highest point in its trajectory. As a result, the SpaceSeeds are pushed out. The second unit is released 10 seconds following the first one's deployment.

\subsubsection{SpaceSeed}
\begin{figure*}
    \centering
    \includegraphics[width=\linewidth]{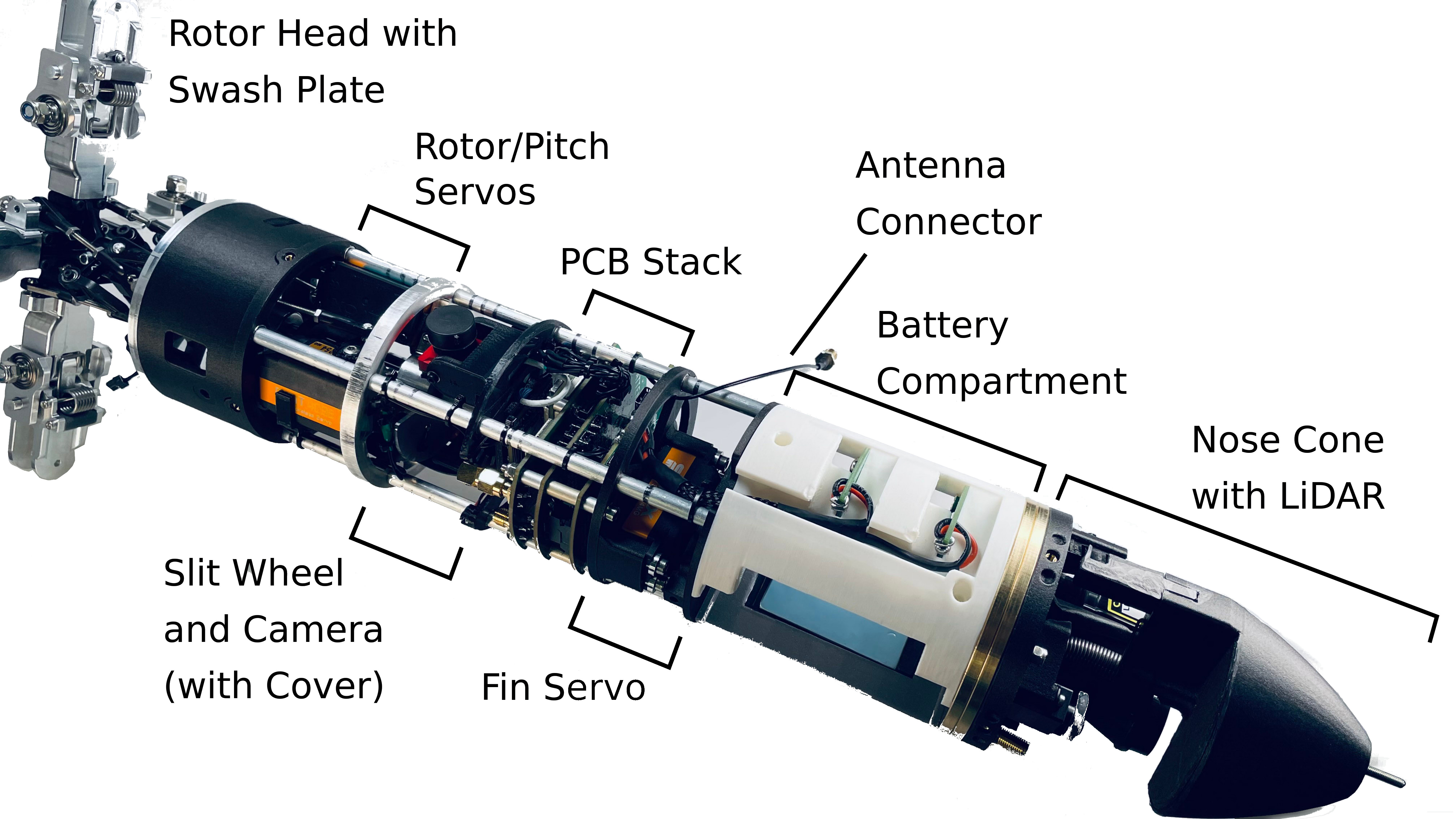}
    \caption{SpaceSeed without hull, annotated sections.}
    \label{fig:seed-no-hull}
\end{figure*}

The two \glspl{ffu} in the ejector each have a mass of \qty{2.6}{\kilogram}. The outer diameter of each units' hull is \qty{85}{\milli\meter}.
The internal structure of the SpaceSeeds consists of a skeletal structure made of four threaded steel rods running along the entire length of the unit, connecting several circular surfaces to which all internal components are mounted. \autoref{fig:seed-no-hull} shows the unit without its hull.

\paragraph{Control Surfaces}
The aerodynamically important parts of the \glspl{ffu} are the two control surfaces: the rotor and the fins.
The latter are attached to the body of the SpaceSeed, which is rotationally decoupled from the rotor and remains at a low rotation rate during flight.
Body rotation is aerodynamically controlled by these two V-shaped fins, stabilizing the \glspl{ffu} and countering rotational moments along the length axis which are created by friction with the rotor.
They are fixed at the top and a servo at their lower fixation allows changing the angle for minimization of the body rotation.

The rotor consists of an off-the-shelf RC helicopter head with four off-the-shelf \qty{470}{\mm} RC rotor blades connected via a tilt mechanism.
It allows folding the blades so that the whole assembly fits into the ejector under the rocket nosecone.
After the ejection, the rotor get deployed due to springs integrated in the foldable joints.
The joints lock in the deployed configuration after ejection, preventing the blades from folding again.
The rotor head's swash plate would mechanically allow cyclic pitch control via three servos placed below it --- however all servos are given common commands and therefore only collective pitch control is used in this stage of the design.

\paragraph{Sensor Accommodations}
The structure also contains accommodations for some sensors.
At the bottom end of the SpaceSeed a LIDAR is hidden behind an inner nose cone for the majority of the flight.
When the \gls{ffu} comes close to the surface, a spring mechanism that gets unlocked by software rotationally removes the cover.
During high-velocity flight the LIDAR thus is protected from particles.
This also minimizes the aerodynamic impact of the opening.
Accurate measurement of the rotor rotation rate is required by the deployed control algorithms (see \ref{sec:control}).
A conventional slit wheel is connected to the rotor axle with two optical sensors determining the rotation rate and direction.
Two optical sensors allow determining the rotation rate and direction (see also \ref{sec:elec-tacho}).
A camera looking radially outwards is also placed next to this assembly.
It is purely used for post-flight analysis and does not serve any purpose within the control scheme.

\subsection{Electronics and Power Management}
This type of scientific experiment introduces complex challenges to the electronic subsystems, spanning from mechanical durability to power management, sensor fail-safes and subsystem communication.
Building on the insights from the \D1 mission \cite{IACD1}, several innovative solutions have been introduced to address these challenges, including redundant communication pathways, mechanical isolation of PCBs, and fail-safe power source selection.
A detailed discussion is available in \cite{wolf2022electronics}.

\subsubsection{Power}
The chosen battery chemistry is \ch{Li/SO2} because of its key advantages: Exceptionally long shelf life with a self-discharge of less than \SI{3}{\percent \per year}, high energy density of \SI{0.74}{\mega\joule\per\kilogram} and a wide operating temperature from \SI{-60}{\celsius} to \SI{70}{\celsius} make it an excellent choice. The non-flammable nature complies with safety regulations which could not have been met with conventional Lithium based cells.

These were employed in two parallel strings with a custom power source multiplexer which would handle load balancing and could switch in case of a short or open circuit in one of the strings.
Additionally, this circuit handles the transition from the rocket power supply to internal batteries and manage the radio-silence and kill-switch features as required by launch site safety regulations.
These are extra layers of operational security to ensure no flow of current in the critical phase of rocket motor integration and during transport.

\ch{Li/SO2} has the disadvantage of a particularly strong passivation layer, which builds up after long storage and inhibits current flow while present.
This passivation layer is the cause of the low self-discharge and must be removed before use, otherwise it would cause a significant voltage drop during operation.
Therefore, before final integration, the batteries are loaded with increasing currents until they reach an acceptable voltage drop at 150 percent of the highest expected current.
This high margin of error was chosen because the passivation layer has already caused problems on NASA's 2003 Mars Exploration Rover where the voltage drop was much higher than expected after the long journey \cite{Ratnakumar2004}.
Fortunately, we did not encounter such problems.

\subsubsection{Selected Sensors}

\paragraph{Tachometer}
\label{sec:elec-tacho}
Reliable measurement of the rotor's rotational speed is essential for both envelope protection and precise descent and landing controls. To accomplish this, optical switches have been paired with a slit wheel attached to the rotor shaft. This configuration interrupts the light path at a frequency directly proportional to the rotor's rotational speed.
By using the powerful timers of the employed STM processor, the time difference between two pulses could be determined with a resolution of \SI{12}{\nano\second}, which allowed for very precise speed estimation of the rotor.

\paragraph{Barometer}
\label{sec:elec-baro}
Redundancy was achieved through the integration of two barometers, as they are crucial for altitude assessment and consequent selection of the appropriate control algorithm.
To depend solely on the \gls{gnss} system was considered to risky.
The \gls{cop} sensor excels in detecting lower minimum pressures (rated for 1\% down to \SI{400}{\pascal}), while the \gls{sbc} sensor offers superior resolution ($\pm$ \SI{100}{\pascal} down to \SI{1000}{\pascal}).

\subsection{Embedded Software}  
The onboard software is separated in three main parts, respectively locations: The \gls{rbc} on the ejector, the \gls{sbc}, and the \gls{cop} on the SpaceSeeds.
Each part is written in C++ and is built on the RODOS embedded operating system and middleware \cite{montenegro2009rodos}.
Using RODOS enables the pseudo-concurrent execution of time-critical tasks, like running the controller, handling of communication,
or interfacing with sensors. 

\subsubsection{Seed Board Computer}
The \gls{sbc} software is responsible for controlling the SpaceSeed's actuators, reading the sensors and storing data in the flash storage.
It also handles telemetry and telecommands of the SpaceSeed. 
The central operations are performed in the main thread in one big loop.
A linear state machine keeps track of the flight state.
The behavior of the system (sensor logging, telemetry, controller etc.) are strictly dependent on the state of this state machine.
There are states for flight preparation, radio silence, rocket flight, ejection, falling and recovery.
Additionally, a debug state is implemented, which can be reached from any state, for testing and debugging purposes.
The core part of the experiment is executed in the falling state.
The first action is always handling sensor reading and subsequent update of state estimation filters.
Based on this, the control output is computed (cf. \ref{sec:control}) and the resulting inputs are sent to the servos affecting the control surfaces.
In every loop iteration all relevant data is collected and stored redundantly on two flash memory chips.
This includes all sensor data, the current state of the state machine as well as several status and error codes. 
After ejection, the position is periodically transmitted via the Iridium module. 
In the recovery state, the SpaceSeeds additionally send their position via LoRa to assist the recovery crew in locating the \glspl{ffu}.

\subsubsection{Seed Co-Processor}
The \gls{cop} is an additional processor on the \gls{ffu} located on a different PCB than the \gls{sbc}.
Its purpose is to control the power supply, measure current and voltage and to control the battery heater. 
The \gls{cop} is connected via UART to the \gls{sbc}.

\subsubsection{Rocket Board Computer}
The software of the \gls{rbc} located on the ejector is responsible for communication with the rocket's telemetry and telecommand system.
It also provides a bridge to the SpaceSeeds and turns them off and on when needed, e.g. during radio silence.
Another task is to trigger the camera mounted on the ejector. 
The software's behavior is also defined by a state machine, similar to the Seed Board Computers.
There are states for flight preparation, radio silence, the rocket flight, ejection, and after the ejection.
As with the \gls{sbc}, there is also a debug state.

\subsubsection{Surviving Reboots}
Much attention has been devoted to being able to continue the experiment after an unexpected reboot.
A sudden acceleration might cause the servos to draw more current than normal, resulting in a voltage drop and subsequent reset of one or more CPUs due to the integrated brown out protection.
To counteract this, one EEPROM chip is used to store the current state and time as well as the last written flash page at regular intervals.
It has a higher write endurance and a faster rewrite speed per storage location compared to the flash memory used for logging.
During the boot sequence, the system tries to restore its state from the last saved one stored on the EEPROM.
An additional safety margin is added as well, as e.g. some flash pages may have been written to in the time between the last store on the EEPROM and the reset.

\subsubsection{Telecommands}
The system also supports a great number of telecommands, which were especially important for configuring flight parameters affected by meteorologic measurements on the day of launch.
They are also used to configure the system for either a dry pre-launch test without ejection or the flight with ejection.
During these dry pre-launch tests, activating the servos --- as happens normally after ejection --- while the SpaceSeeds were still in the ejector tubes would have blocked the servos, causing high power draw and possibly thermal destruction of these components.

The commands were interpreted by the console subsystem of our software stack.
It allowed dynamically executing predefined commands as one might know from a command prompt.
It accepts ASCII input, with command name and parameters in human-readable form separated by spaces.
Using this subsystem for telecommands results in larger packets due to the need to transport human-readable text.
However, it increases the flexibility of the system, as fewer components need to be changed when adding a new command.
This simplicity decreases the failure probability.

\subsection{Ground Station Software}  
\label{sec:ground-station}

The Ground Station / Mission Control software is based on our own open-source framework, Telestion. Its source code (except for the components pre-processing the Iridium data) is available on GitHub\footnote{https://github.com/wuespace/telestion-project-daedalus2}. Two different modes are used to support the two main mission stages: before and after the launch of the rocket.

\begin{figure}
    \centering
    \includegraphics[width=0.96\linewidth]{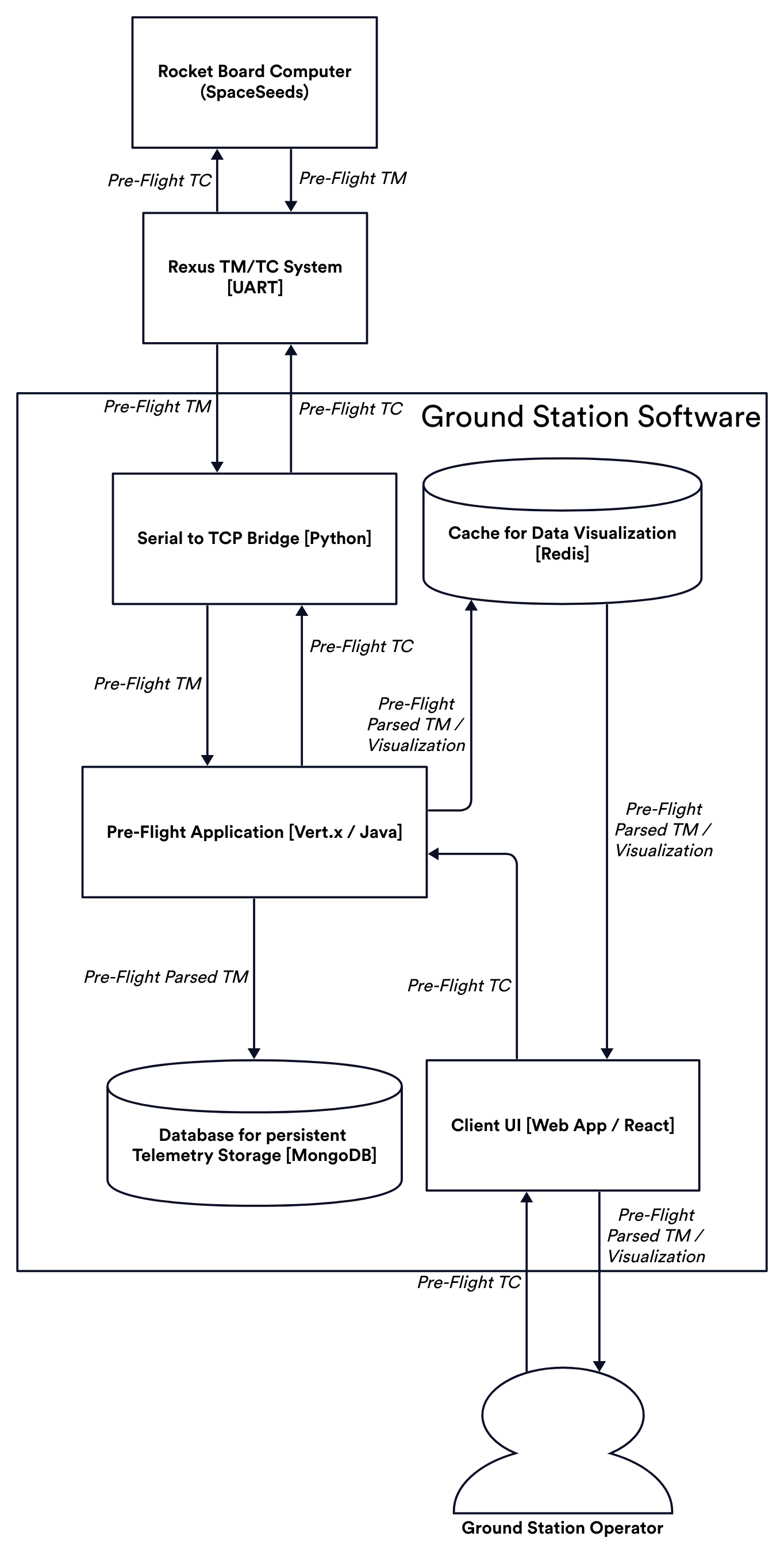}
    \caption{Daedalus 2 ground station software structure during the pre-flight stage.}
    \label{fig:gs-preflight}
\end{figure}

\paragraph{Before launch} This mode enables the operator to monitor the state of the SpaceSeeds in real-time as well as allowing for telecommands to the different components, for example, but not limited to, entering radio silence modes. Due to the requirements of the Esrange Space Center, this had to be deployed in a separate network that had to be physically isolated from any remote access. You can see the internal software architecture in \autoref{fig:gs-preflight}.

\paragraph{After launch} This mode saves any telemetry received through the Iridium Short Burst Data (SBD) service via TCP. This enables operators to analyze data relevant to recovery in real-time. Due to the remote-server-based architecture, both, the flight team at Esrange and a remote support team in Würzburg were able to access the same data through the same interface simultaneously. Authorization across these sites was handled through a virtual private network (VPN). The architecture was mostly shared with the pre-flight structure shown in \ref{fig:gs-preflight}, but without any capability to send telecommands, and with an \textit{Elixir-based Iridium SBD Data Receiver} instead of the \textit{Serial to TCP Bridge}.

The Telestion framework supports various deployment setups using the same code but with different configuration files. As a result, most of the code was consistent across both deployments, simplifying maintenance.
The selected system design permitted the Java-based ``Application'' (referring to the backend) to run alongside its associated Redis cache layer and MongoDB database within a docker-compose environment, as all interfaces into and out of that system were based on TCP connections.

Throughout the mission, all operations of the Ground Station Software ran smoothly, with operators reporting no issues.

\subsection{Control}  
\label{sec:control}
Numerous control methods were evaluated during development and testing.
NMPCs \cite{dalamagkidis2009autonomous}, Neural Networks \cite{nonami2010autonomous} and PIDs were amongst the tested strategies.
Collective pitch and actuation of the body fins are the outputs.
They are computed from the rotor rotation rate, multiple altitude and vertical speed readings.

The flight phase were already introduced in \ref{sec:profile}.
Each phase has different control goals and constraints, so different control strategies are used to target the changing requirements.
They are outlined here in short and are explained in more detail in a control-specific paper \cite{riegler2022modeling}.

\begin{itemize}
    \item Fast Descent Control
    \begin{itemize}
        \item Control Goal: 500~\acrshort{rpm} rotor rotation rate.
        \item Controller: Analytically tuned PID through linearization.
    \end{itemize}
    \item Glide Control
    \begin{itemize}
        \item Control Goal: 2000~\acrshort{rpm} rotor rotation rate.
        \item Controller: Genetically tuned PID.
    \end{itemize}
    \item Landing Control
    \begin{itemize}
        \item Control Goal: $\SI{0.5}{\meter\per\second}$ touchdown speed.
        \item Controller: Genetically tuned PID following a velocity ramp function.
    \end{itemize}
\end{itemize}

\section{Flight}
\label{sec:flight}



The flight campaign was performed on April 1st 2023 in the early morning hours in Esrange.
After launch, the SpaceSeeds were deployed from the rocket, confirmed by data and visually from the camera feed.
During flight, both vehicles struggled to connect with the Iridium network.
We received only few pings from each \gls{ffu} during the flight.
A valid \gls{gnss} position could not be retrieved from the data.
After the vehicles had landed, we at first only received data from one system.
SpaceSeed A was sending data continuously until it was picked up by the recovery crew.
Further messages from Unit B were only received after A mostly stopped sending, about \qty{90}{\minute} after launch.
The reason for this remains unclear, as no logging was performed after touchdown to save energy for the recovery mechanisms. 

The recovery crew was able to find and recover both SpaceSeeds using a helicopter from the \gls{gnss} positions received after the \glspl{ffu} had landed.
Additionally, use of the RECCO system aided in locating the precise position of the \glspl{ffu}.
Both units had to be dug out of the snow, only a hole was visible from the surface.
The rotor was not attached anymore and could not be recovered.
Both SpaceSeeds were missing their main rotor.
The vehicles' state right after the recovery can be seen in \autoref{fig:recovered-seeds}.

When we received the units from the recovery crew, the \glspl{sbc} on the \glspl{ffu} were still running nominally.
The debug connection was intact and thus used for initial download of logged flight data.
All sensor data and video files were recovered from the vehicles, which aided the failure analysis.

\subsection{Loss of the Rotor}  
\begin{figure}
    \centering
    \includegraphics[width=0.95\linewidth]{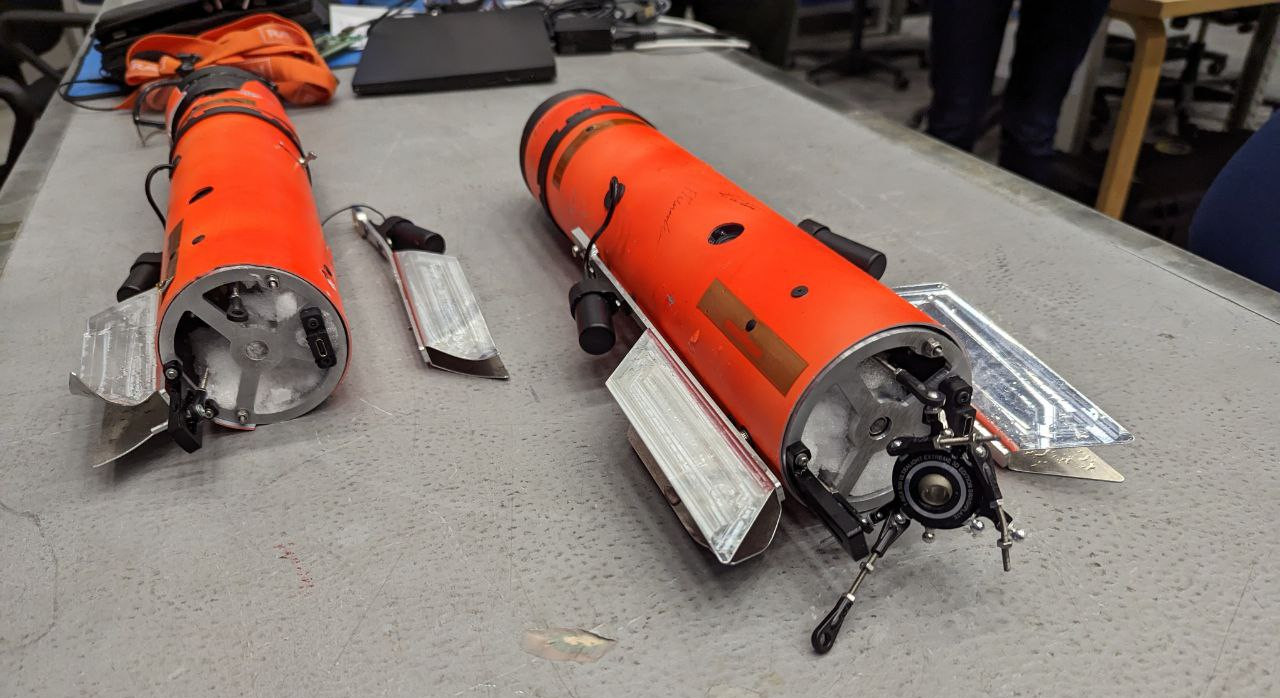}
    \caption{SpaceSeeds on the workbench right after recovery with the main rotor missing.}
    \label{fig:recovered-seeds}
\end{figure}

During the flight, both vehicles lost almost the entire rotor head.
The SpaceSeeds, without a rotor head, are depicted in \autoref{fig:recovered-seeds}.
One clear element is that the rotor axle broke at a phase where it interfaced with a ball bearing.
When analyzing the remainder of the axle, it became clear that the rotor broke by bending loads.

Both SpaceSeeds featured cameras that also had the rotor blades partially in view.
During the descent, the vehicle was comparatively unstable, and the fins did not appear to stabilize it.
This led to increased loads that the system wasn't designed for. 
It was visible that the rotor blades suddenly stopped spinning and reoriented chaotically.
None of the rotor blades appeared to be broken.
However, on some blades, the top side was visible.
Assuming the blade is not broken, this only occurs when the struts that connect the rotor blade to the swash disk fail.
Ultimately, in both videos, the rotor head can be seen separating from the vehicle, confirming that this happened early in the flight.

This leads to either one of two possible scenarios.
\paragraph{The axle broke first}
Once the axle was broken, the rotor head moved chaotically and was only held by arms connecting the blades to the swash-disk.
Once enough of these arms had failed, the rotor head detached.

\paragraph{The swash-disk arms broke first}
This led to chaotic behavior of the rotor blades, moving freely in the air stream.
Subsequently, this contributed to the vehicle's instability, increasing the loads on the axle.
Ultimately, the axle broke due to bending forces, and the rotor head detached.

Whether either scenario is correct will remain unclear. 
Nevertheless, the outcome is quite similar.
The chosen rotor head was not suitable for the loads it faced.
A stronger rotor, both concerning the axle and connection of blade hinges and swash-disk, is likely to resolve the problem.

\subsection{Data Analysis}  
The recovered sensor logs indicate that no reboot occurred during the flight.

\begin{figure}
    \centering
    \includegraphics[width=0.95\linewidth]{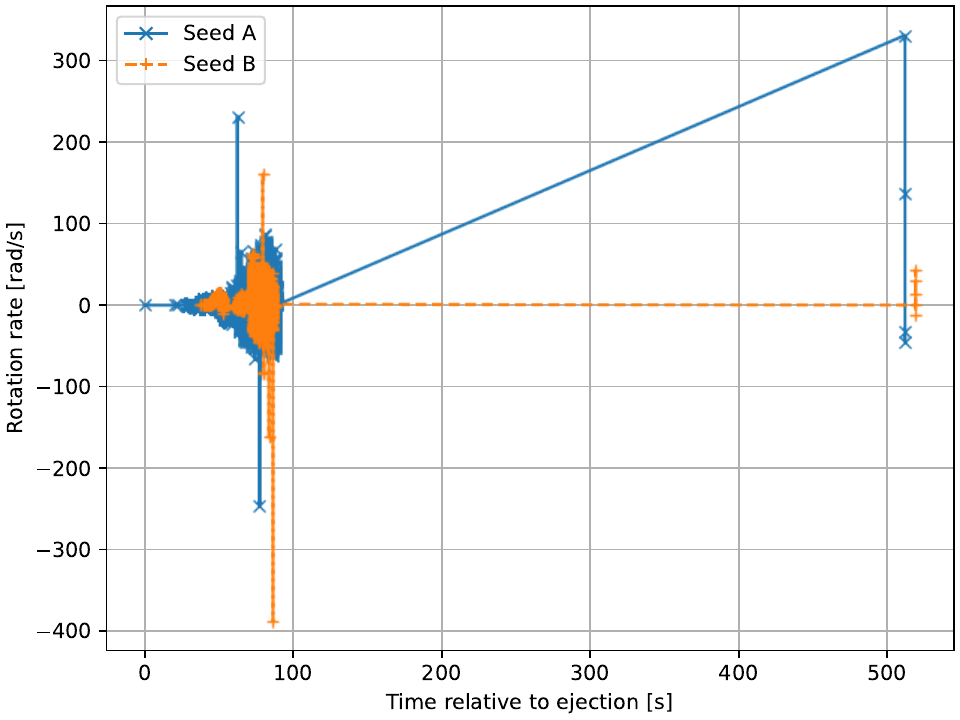}
    \caption{Rotor Rotation Rate of both SpaceSeeds, relative to the ejection of each unit at $t=\qty{0}{\s}$.}
    \label{fig:rotor-rot-rate}
\end{figure}

\autoref{fig:rotor-rot-rate} shows the rotor rotation rate of each \gls{ffu}.
Available measurement points are marked with a cross (unit A) or a plus sign (unit B) when available.
Measurements are not necessarily available for each loop iteration:
When no edges have been encountered since the timer was last triggered a \emph{not available} marker was written to storage instead of \qty{0}{\radian\per\s}.
This was done to not submit zero to later processing stages when the rotor was in fact only moving very slowly.
Long stretches without data points must therefore be manually interpreted as \emph{no rotation}.

As can be seen from the plot, both rotors started spinning up about \qty{30}{\s} (unit A) to \qty{45}{\s} (unit B) after ejection.
The loss of the rotors happens much closer together though, with SpaceSeed B's last measurement point occurring approximately \qty{88}{\s} after ejection and unit A following just \qty{1.5}{\s} later.
The spikes later at about \qtyrange{510}{520}{\s} occur together with spikes of the accelerometer and gyroscope magnitude before both sensors settle to their resting values.
It is assumed that the impact shock is the main cause for the measured high rotation rate, as the \glspl{ffu} impacted with about \num{110} (unit B) to \qty{140}{\km\per\hour} (unit A) according to barometer data.

\begin{figure}
    \centering
    \includegraphics[width=0.95\linewidth]{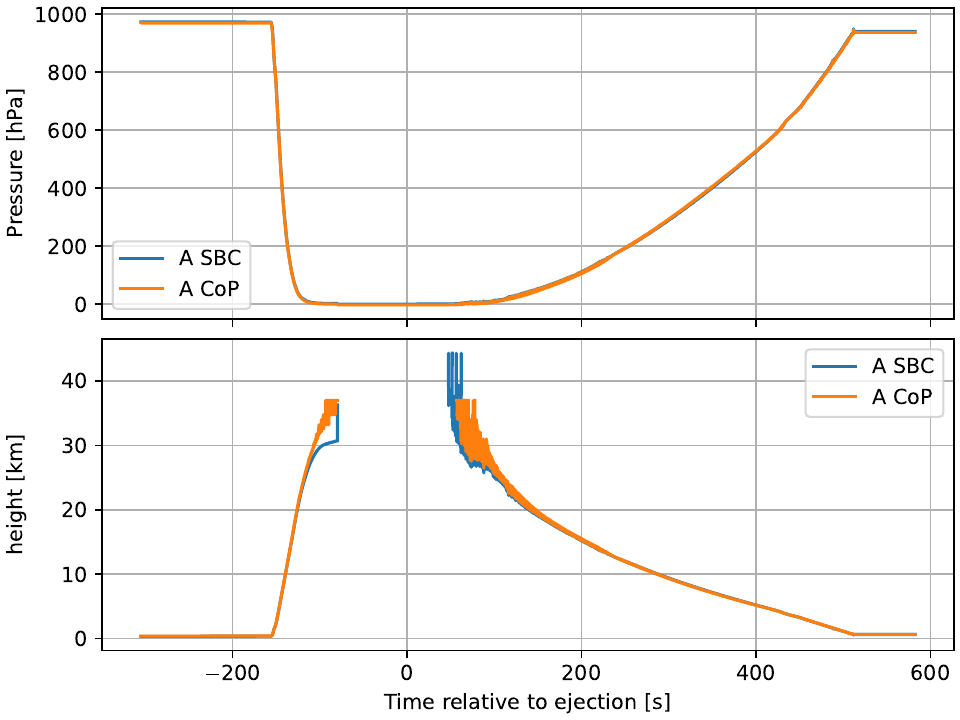}
    \caption{Barometer measurements and derived height from SpaceSeed A}
    \label{fig:baro-plot}
\end{figure}

At altitudes below \qty{30}{\km}, both barometers (see \ref{sec:elec-baro}) demonstrated highly comparable performance, as shown in \autoref{fig:baro-plot}.

Both SpaceSeeds also did not get a \gls{gnss} lock during the descent.
We assume that the tumbling motion, as evident from the videos and sensor data, was the main reason for this.

\section{Team Management}  
\label{sec:team}

A considerable, often underappreciated task is team management.
Managing a team of volunteering students is a special challenge as some unique circumstances have to be accounted for.
Most team members are ``learning on the job'' and doing tasks they have never done before.
Since a student project is almost always on a volunteer basis, passion has to replace obligation completely.
Typical workers might push through things they don't like because of their contract. 
Pushing a team of students through extremely rough patches demands passion within the team.
\D2 had to go through extraordinary hardships during the project run time.
In this section, we want to introduce how these were solved and how our team was generally organized.

\subsection{Team}
In the previous sections, it became clear that this project faced high technical complexity.
To overcome this complexity, we needed a substantial team. 
In total, 40 team members have contributed to this project.
They were organized into subteams, which had tasks assigned to them.

\begin{itemize}
    \item Software
    \item Ground Station Software
    \item Simulation \& Control
    \item Electronics
    \item Mechanics
    \item PR \& Outreach
    \item Finance \& Sponsoring
\end{itemize}

Each team was headed by one or two team-leaders who directly reported to the project leaders.
This reporting was done weekly in a mandatory meeting for project- and team-leads. 
Everyone else was welcome as well.
The teams themselves had at least one weekly meeting.
The meetings were held via Discord to make them more accessible and convenient for team members.
However, before COVID, meetings were regularly held on campus as well.
Teams sometimes scheduled meetings with other teams if interfaces or common issues had to be discussed.
Overall, the organization was comparatively autonomous and self-governed.

\subsection{Project Timeline}
\label{sec:project-timeline}
\begin{figure*}
    \centering
    \includegraphics[width=\textwidth]{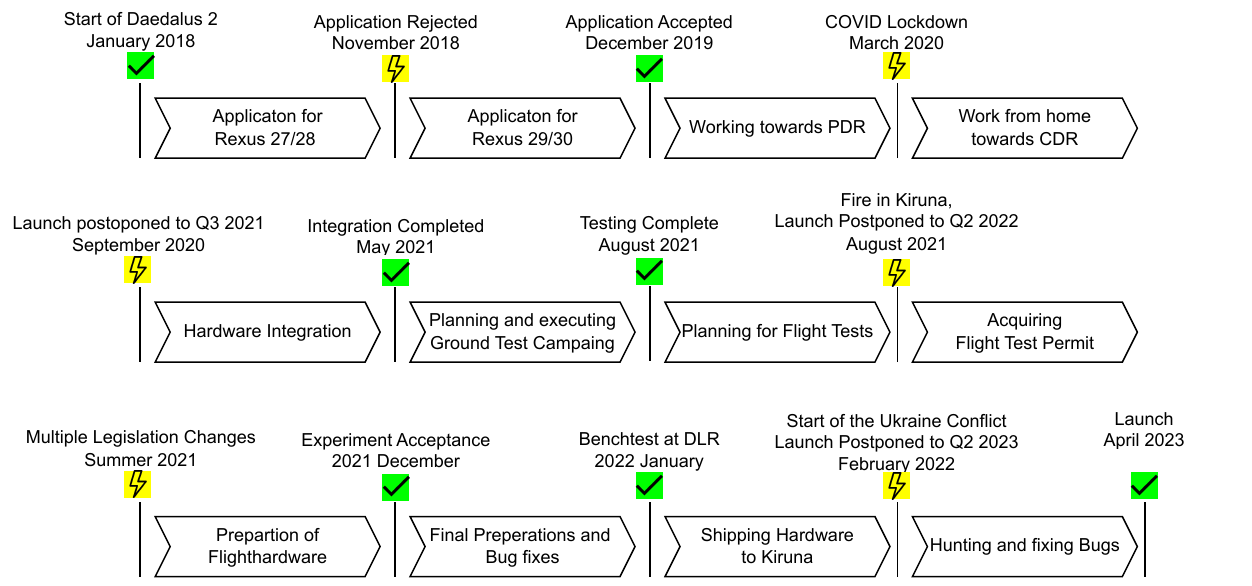}
    \caption{Event-based timeline illustrating milestones and unforeseen events having a major influence on the project's progression.}
    \label{fig:timeline}
\end{figure*}

Contrary to other REXUS projects, \D2 was faced with a unique timeline.
From application to launch, including some time to prepare the applications, it usually takes about two years.
In this case, due to force majeure, it took just over five years to complete this project.
The core events that postponed our launch are listed below.
A more detailed event-based timeline is depicted in \autoref{fig:timeline}.

\begin{itemize}
    \item Rejection due to ongoing Project \D1 (postponed due to malfunction of REXUS 24 \cite{volk2019rexus}).
    \item Outbreak of the COVID-19 epidemic in 2020.
    \item Fire in Esrange destroying key infrastructure for launch in 2021 \cite{esrange}.
    \item Start of the Russo-Ukrainian war, days before the scheduled launch campaign in 2022.
\end{itemize}

These stepping stones had major effects on team cohesion.
Naturally, some students finished their studies during that time and were not available anymore.
Thus, new students had to be introduced to the project, mainly positively affecting the project.
Furthermore, many more team-building measures had to be implemented.
This was especially challenging in times of COVID.

\subsection{Workflow}
The work was scheduled in milestones and sub-milestones and managed with Trello.
The tasks were split into sprint-like time frames and finished to reach certain goals.
Thereby, the teams could track their progress and see other teams progress.
Trello also allowed interaction on certain work packages and sharing files in the needed context.
Thus, a Scrum-like workflow was established.
Regular tests worked as anchors and "reality-checks" to see whether goals were achieved and if interfaces worked as intended.


\section{Conclusion}
The paper presented the \D2 mission designed that was flown with the goal to demonstrate autorotation and the results.
The stated mission goal must be classified as failure, as the rotor was lost soon after ejection and therefore autorotation could not be demonstrated.
Nevertheless, the team members gained valuable experience during this project as part of their education.
Additionally, we were able to show that the electronics survived the harsh conditions and the impact from free fall in the snow.
The software also worked as expected and no anomalies occurred.
As the same mechanical failure occurred on both systems, it should be possible to address this shortcoming in a new design iteration.
This failure does therefore not imply that autorotation as an \gls{edl} mechanism in general is not a viable strategy.
Since this is a student project with no immediate successor, we wish the best of luck to any project in the future trying a similar setup.

\section{Acknowledgements}
The REXUS/BEXUS program is realized under a bilateral Agency Agreement between the German Aerospace Center (DLR) and the Swedish National Space Agency (SNSA). The Swedish share of the payload has been made available to students from other European countries through the collaboration with the European Space Agency (ESA).
Experts from DLR, SSC, ZARM and ESA provide technical support to the student teams throughout the project. EuroLaunch, the cooperation between the Esrange Space Center of SSC and the Mobile Rocket Base (MORABA) of DLR, is responsible for the campaign management and operations of the launch vehicles.
We are very grateful for their support.

We are also very grateful to Prof. Hakan Kayal and Prof. Sergio Montenegro with their teams. This project would not have been possible without the support of the computer science chair VII at the University of Würzburg.

We would like to extend additional thanks to our sponsors, 
Mouser Electronics,
Carbon-Team,
Hacker Brushless Motors,
MathWorks,
Atomstreet,
Siemens,
u-blox,
VDI,
Breunig Aerospace,
Airbus,
TASKING,
ANSYS,
Hirsch KG,
PCB Arts,
Vogel Stiftung,
MÄDLER,
Würth Elektronik,
PartsBox,
Molex,
LRT Automotive,
Zentrum für Telematik,
Iridium,
Telespazio, and
Artic.

\bibliography{bibliography}

\end{document}

%% file: abbreviations.tex
\newacronym{cop}{CoP}{seed co-processor}

\newacronym{edl}{EDL}{entry, descent and landing}

\newacronym{ffu}{FFU}{free-falling unit}

\newacronym{gnss}{GNSS}{global navigation satellite system}

\newacronym{rbc}{RBC}{rocket board computer}
\newacronym{rpm}{RPM}{rotations per minute}

\newacronym{sbc}{SBC}{seed board computer}

